%% file: main.tex
\documentclass[a4paper, 10 pt, conference]{ieeeconf}

\IEEEoverridecommandlockouts                              

\usepackage{graphics} 
\usepackage{epsfig} 
\usepackage{mathptmx} 
\usepackage{times} 
\usepackage{amsmath} 
\usepackage{amssymb}  

\usepackage{subcaption} 
\usepackage{listings}
\usepackage{algorithm}
\usepackage{algorithmic}

\usepackage{siunitx}

\usepackage{xcolor}
\usepackage{svg}

\usepackage{hyperref}

\usepackage[font=small,labelfont=bf]{caption} 

\usepackage[
backend=biber,
style=ieee,
sorting=none,
]{biblatex}
\title{
 From Obstacle Avoidance To Motion Learning \\
 Using Local Rotation of Dynamical Systems
 \vspace{-0.1em}}
\author{Lukas Huber$^{1}$
}

\begin{document}
\newcommand{\vect}[1]{\mathbf{#1}}
\newcommand{\matr}[1]{\mathbf{#1}}

\newcommand{\angs}[1]{\hat{\mathbf{#1}}}

\newcommand{\dotprod}[2]{\left\langle {#1}, \, {#2} \right\rangle}
\newcommand{\normdotprod}[2]{\frac{\left\langle #1, \, #2 \right\rangle}{\| #1 \| \, \| #2 \|}}

\newcommand{\lemma}[2]{
  \noindent\textbf{Lemma  \thesection. \lcounter / #1:}
  \textit{#2}
}

\newcommand{\theorem}[2]{
  \noindent\textbf{Theorem #1: } 
  \textit{#2}
}

\maketitle
\begin{abstract}
  In robotics motion is often described from an
external perspective, i.e., we give information on the obstacle
motion in a mathematical manner with respect to a specific
(often inertial) reference frame. In the current work, we propose to describe the robotic motion with respect to the robot itself.
Similar to how we give instructions to each other (“go straight,
and then after xxx meters move left, and then turn left.”), we
give the instructions to a robot as a relative
rotation. \\
We first introduce an obstacle avoidance framework that allows
avoiding star-shaped obstacles while trying to stay close to an initial (linear or nonlinear) dynamical system. The framework of the local rotation is extended to motion learning. Automated clustering defines regions of local stability, for which the precise dynamics are individually learned. \\
The framework has been applied to the LASA-handwriting dataset and shows promising results.
\end{abstract}


\input{input/introduction}

\input{input/literature}

\input{input/rotationmodulation}

\input{input/kmeans_obstacle}

\input{input/conclusion}


\renewcommand*{\bibfont}{\footnotesize}
\printbibliography

\end{document}

%% file: input/introduction.tex
\section{Introduction} \label{sec:introduction}


Motion learning and programming by demonstration have seen large development in recent years, thanks to progress in machine learning algorithms, improved computational capacities but also the large availability of data. 
In this work, we introduce a framework that allows the combination of the two approaches. This not only allows the exploitation of synergies between learning and avoiding but also allows the division of the task into globally learned motion and local obstacle avoidance.

\subsection{Properties}
The learned dynamics have the following properties:
\begin{itemize}
\item continuously differentiable ($C^1$ smooth). 
\item globally asymptotically stable. 
\item the learning can be applied to any (non-crossing) trajectory. Specifically, motions that lead away from the attractor in the radial direction,
  i.e.,
  \begin{equation}
    \langle f(\xi), \xi - \xi^a \rangle / ( \|f(\xi) \| \, \|\xi - \xi^a \|)= (-1)
  \end{equation}
  and spiraling motion
\end{itemize}


Additional advantages of the systems are:
\begin{itemize}
\item The motion can be constrained to a region of influence using a radius $R^{\mathrm{kmeans}}$ to ensure proximity to the initial data points, i.e., creating an invariant set around the \textit{known} environment.
\item The motion learning can be combined with avoidance algorithms to ensure convergence towards the attractor around (locally \textit{star-shaped}) obstacles for the constraint region.
\end{itemize}

\begin{figure}[t]
    \centering 
    \includegraphics[width=0.5\columnwidth]{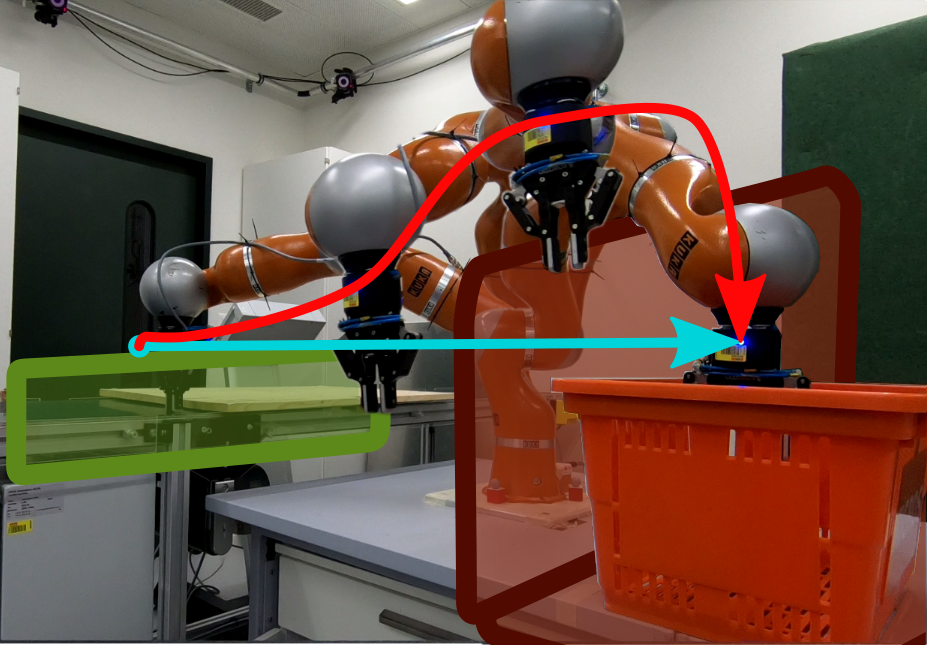}
    \caption{Robotic arm avoiding obstacles while following a previously learned dynamical system \cite{huber2019avoidance}}.
    \label{fig:avoiding_learned_trajectory}
\end{figure}

%% file: input/literature.tex
\section{Literature}


A popular method of learning from demonstration is SEDS \cite{khansari2012dynamical}. It ensures stability by using a quadratic Lyapunov function,  this comes with the cost of low accuracy for non-monotonic trajectories (temporarily moving away from the attractor).
A parametrized quadratic Lyapunov function offer more flexibility, but still struggle to approximate non-linear trajectories \cite{figueroa2018physically}.

More complex Lyapunov functions can be automatically learned to ensure the stability of the final control  \cite{khansari2014learning}. This has been further extended to additionally learn the control parameters of the motion \cite{khansari2017learning, figueroa2022locally}. A weighted sum of asymmetric quadratic functions (WSAQF) is used to obtain the final Lyapunov candidate, which limits the motion to not be able to move in radial direction away from the attractor.



Diffeomorphic matching of an initial linear trajectory with the learned trajectory (highly nonlinear) \cite{neumann2015learning, rana2020euclideanizing}. However, the  work cannot give any guarantees on the convergence of the matching. 

Dynamic movement primitives introduce time-dependent parameters which ensure the convergence in finite time towards the goal \cite{ijspeert2013dynamical}. However, this leads to motion increasingly deviating from the learned motion with increasing time.

%% file: input/rotationmodulation.tex
\section{Obstacle Avoidance through Rotation} \label{sec:rotational_avoidance}

\subsection{Preliminaries}
We restate concepts developed in \cite{huber2019avoidance, huber2022avoiding}.  The most direct dynamics towards the attractor is a linear dynamical system of the form:
\begin{equation}
  \vect f(\xi) = - k (\xi - \xi^a)
\end{equation}
where $k \in \mathbb{R}$ is a scaling parameter. 

Real-time obstacle avoidance is obtained by applying a dynamic modulation matrix to a dynamical system $\vect f(\xi)$:
\begin{equation}
  \dot{\xi} = \matr{M}(\xi) \vect f(\xi)
  \qquad \text{with} \quad
  \matr{M} ( \xi) = \matr{E}(\xi) \matr D(\xi) \matr{E}(\xi)^{-1} \label{eq:modDS_app}
\end{equation}
The modulation matrix is composed of the basis matrix:
\begin{equation}
  \matr {E} (\xi) =
  \left[ {\vect r }(\xi) \;\; \vect e_1(\xi) \;\; .. \;\; \vect{e}_{d-1}(\xi) \right]
   \label{eq:basisMatr}
\end{equation}
which has the orthonormal tangent vectors $\vect e_i(\xi)$.
The diagonal eigenvalue matrix is given as:
\begin{equation}
  \matr D(\xi) =
  \textbf{diag}
  \left(
    \lambda^r(\xi) ,
    \lambda^e(\xi) , \,
     \hdots ,
     \lambda^e( \xi)
     \right)
  \label{eq:eigVecMatr}
\end{equation}

\subsection{Rotational Based Avoidance}
Any modulated dynamics $\dot \xi$ as described in the previous section and the initial dynamics $\vect f(\xi)$ can be interpreted as a rotation and a stretching, i.e., the application of a rotation matrix $\matr R(\xi)$ and the scalar stretching function $h(\xi)$:
\begin{equation}
    \dot{\xi} = h(\xi) \matr R(\xi)  \; \vect f (\xi)
\end{equation}
as proposed by \cite{kronander2015incremental}. When all functions are smoothly defined, the absence of spurious attractors is ensured.


\subsection{Directional Space Rotation}
\begin{figure}[h]
    \centering 
    \includegraphics[width=0.4\columnwidth]{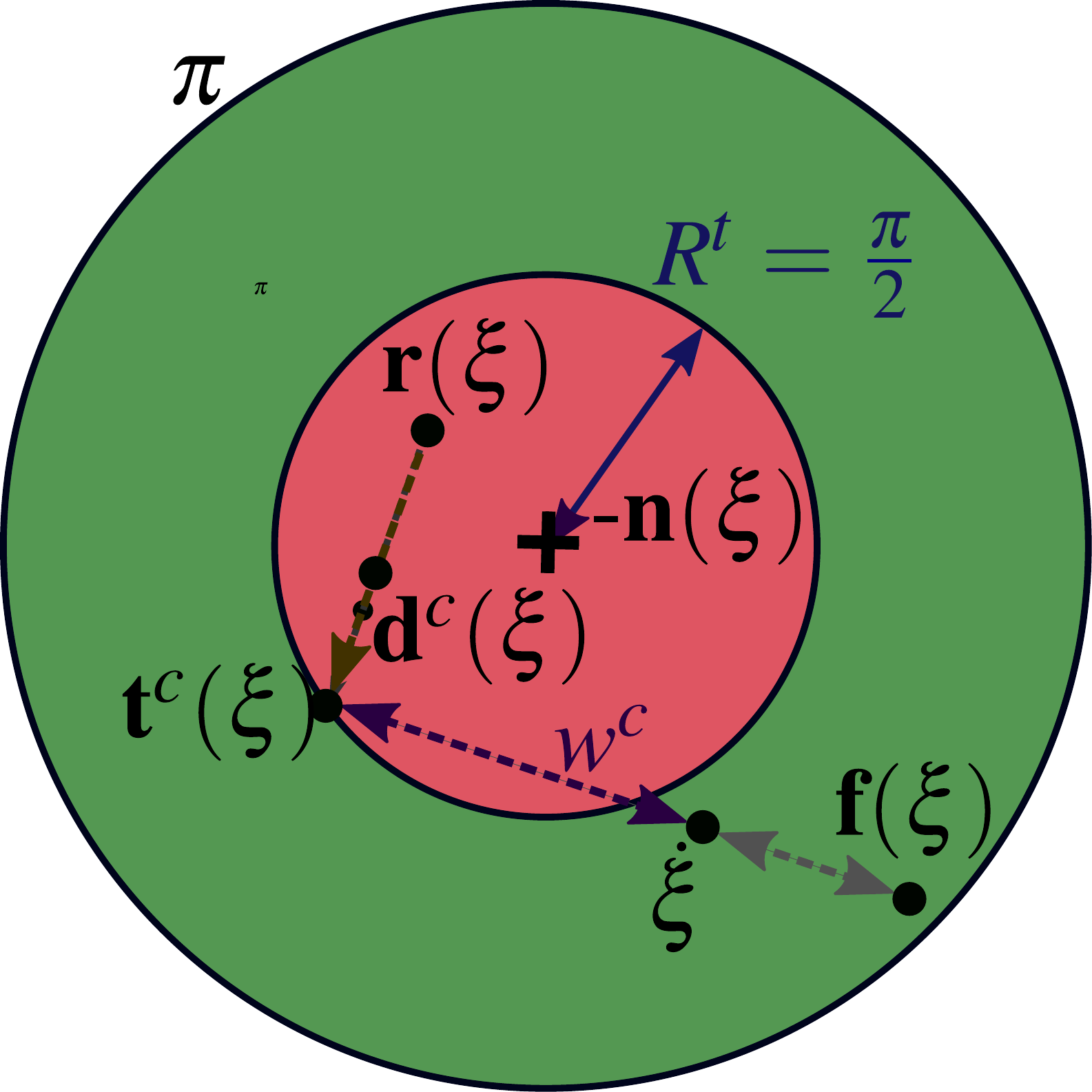}
    \caption{Rotational modulation visualized in the direction spaces. Any direction in the green region (an angle of $> \pi/2$ to the normal $\vect n(\xi$) is collision free. Hence the goal is to \textit{move} the initial velocity $\vect f(\xi)$ into this region.}
    \label{fig:rotation_in_direction_space}
\end{figure}

The algorithm is divided into two steps, (1) finding a tangent direction $\vect t(\xi)$ which guides the velocity around the obstacle, and (2) rotate the initial dynamics $\vect f(\xi)$ towards the tangent (see Fig.~\ref{fig:rotation_in_direction_space}).

\subsection{Rotation of Dynamics}
The dynamics are then \textit{pulled} towards the tangent, the pulling is a function of $\Gamma$, the distance to the obstacle.

\begin{figure}[t]\centering
\begin{subfigure}{.33\columnwidth}
\centering
\includegraphics[width=\textwidth]{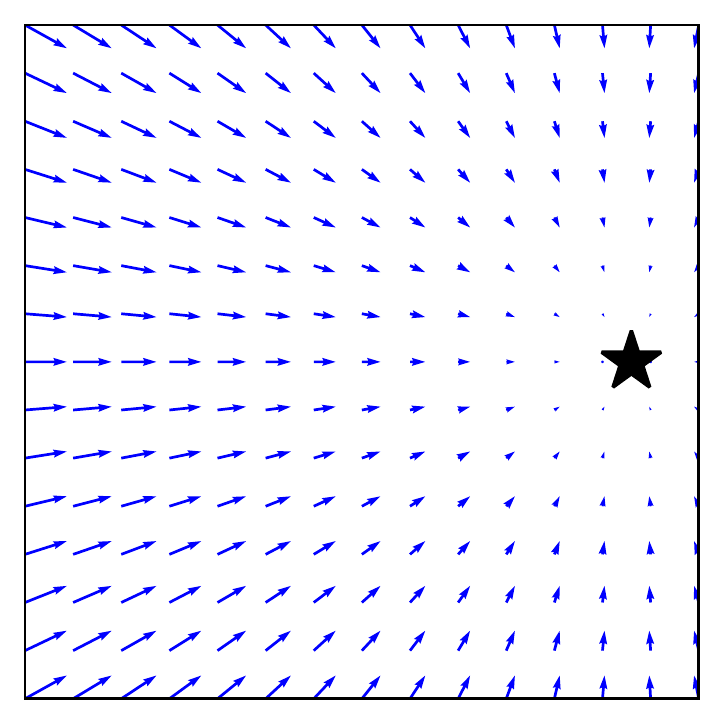}
\caption{Initial dynamics}
\label{fig:linear_vectorfield_initial}
\end{subfigure}%
\begin{subfigure}{.33\columnwidth}
\centering
\includegraphics[width=\textwidth]{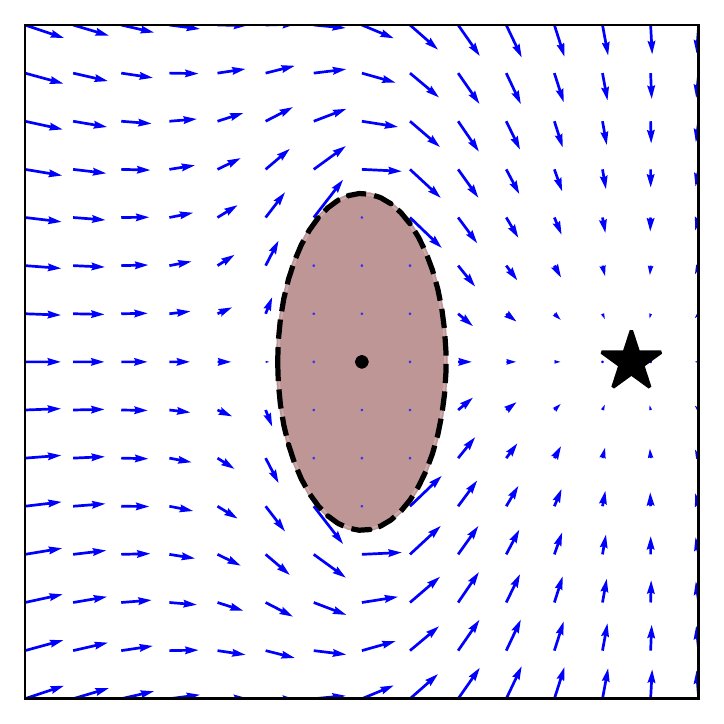}
\caption{Modulated system}
\label{fig:linear_vectorfield_modulated}
\end{subfigure}%
\begin{subfigure}{.33\columnwidth}
\centering
\includegraphics[width=\textwidth]{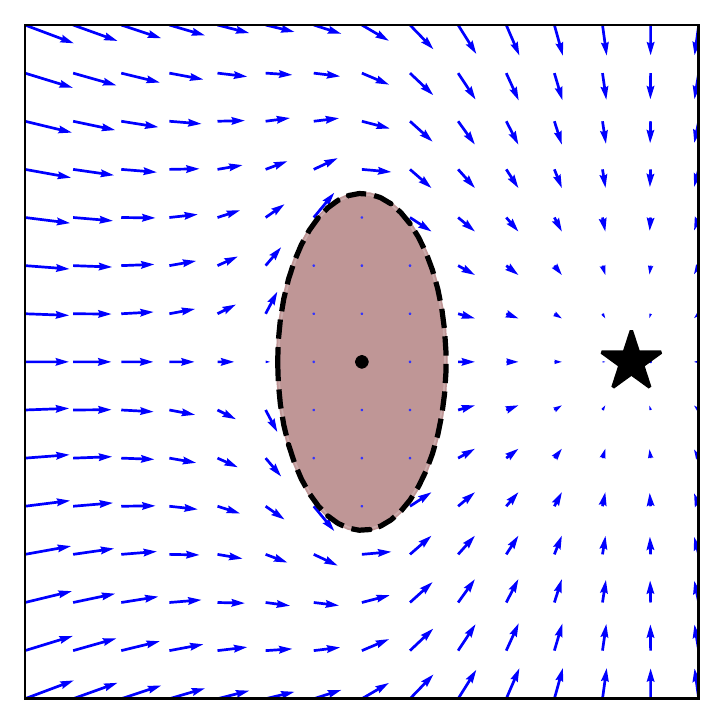}
\caption{Rotated system}
\label{fig:linear_vectorfield_rotated}
\end{subfigure}%
\caption{For linear dynamics \textbf{(a)} the modulation \textbf{(b)} and the ROAM \textbf{(c)} have very similar behavior. However, the ROAM method keeps the velocity identical to the initial velocity while moving around the obstacle.}
\label{fig:comparison_linear_vectorfield}
\end{figure}

The convergence direction $\vect d^c$ can be evaluated as the direction towards the attractor, i.e., we have:
\begin{equation}
  \angs{\dot{\xi}} = w^c \angs{t}^c(\xi) + (1 - w^c) \angs{f}(\xi)
\end{equation}

where the convergence weights are given by
\begin{equation*}
  w^c = \left(w^\Gamma\right) ^{1 / w^{s}}
  \;\; \text{with} \;\;\;
  w^\Gamma = \frac{1}{\Gamma(\xi)}, \;\;\;
  w^s = {\| \angs{r}(\xi) - \angs{d}^c(\xi) \| }^ {c^s}
\end{equation*}
with $c^s \in \mathbb{R}_{>0}$ the smoothness constant weight. The distance weight $w^\Gamma$ ensures decreasing influence with increasing distance $\Gamma(\xi)$, whereas the smoothing weight $w^s$, ensures a smooth continuation where the convergence direction is pointing towards the robot (see Fig.~\ref{fig:comparison_linear_vectorfield}).

\subsection{Inverted Obstacles}
The inversion of the obstacle (moving inside a boundary-hull) is done equivalently to \cite{huber2022avoiding}, i.e. projecting the point outside. It was shown that the distance function of a boundary is evaluated as:
\begin{equation}
  \Gamma^w(\xi) = 1/\Gamma^o = \left( R(\xi)/ \|\xi-\xi^r \| \right)^{2p} \qquad \forall \; \mathbb{R}^d \setminus \xi^r \label{eq:inverse_gamma}%
\end{equation}

Additionally, for the rotational avoidance, we also flip the reference and normal direction (and the resulting decomposition matrix):
\begin{equation}    
    \vect n^b(\xi) = - \vect n(\xi) 
    \qquad
    \vect r^b(\xi) = - \vect r(\xi) 
\end{equation}

\begin{figure}[t]
    \centering 
    \includegraphics[width=1.0\columnwidth]{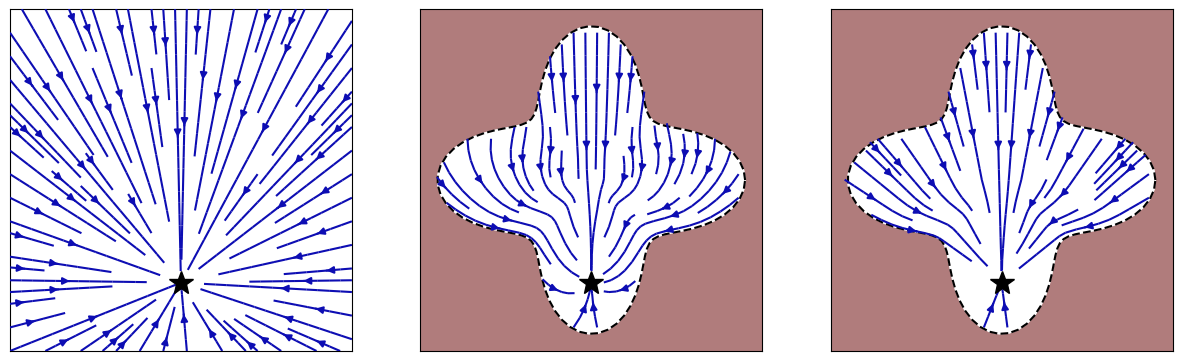}
    \caption{Rotational Modulation Inside a Star-Shaped Hull}
    \label{fig:comparison_inverted_vectorfield}
\end{figure}

Analogously \cite{huber2019avoidance, huber2022avoiding} the algorithm can be extended to multiple obstacles and inverted obstacles, i.e. surrounding hulls (see Fig.~\ref{fig:comparison_inverted_vectorfield})

\begin{figure}[t]\centering
\begin{subfigure}{.33\columnwidth}
\centering
\includegraphics[width=\textwidth]{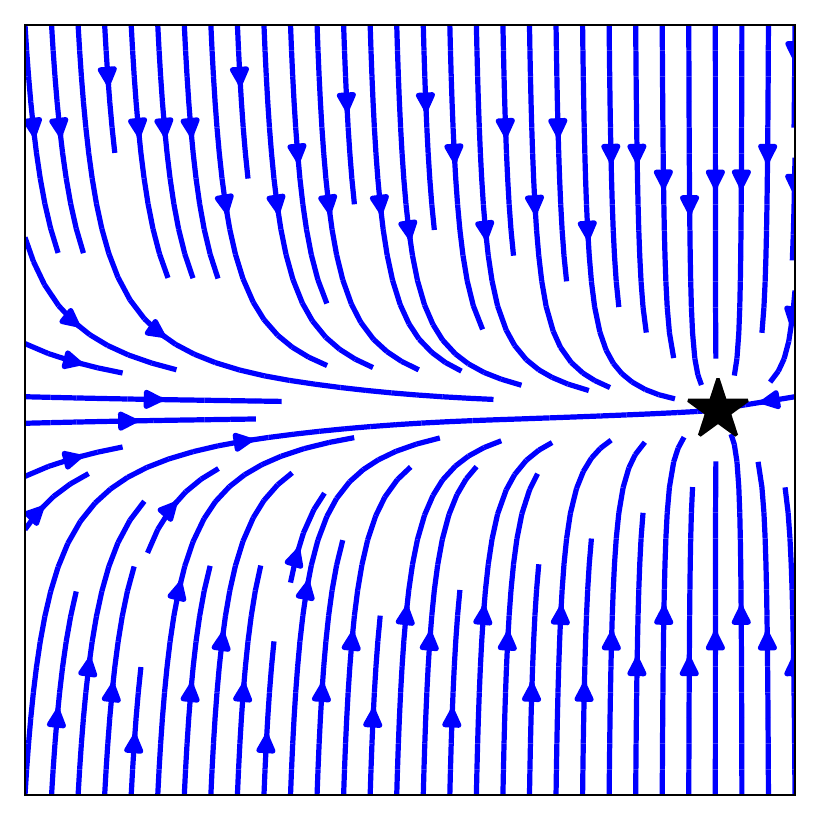}
\caption{Initial dynamics}
\label{fig:nonlinear_vectorfield_initial}
\end{subfigure}%
\begin{subfigure}{.33\columnwidth}
\centering
\includegraphics[width=\textwidth]{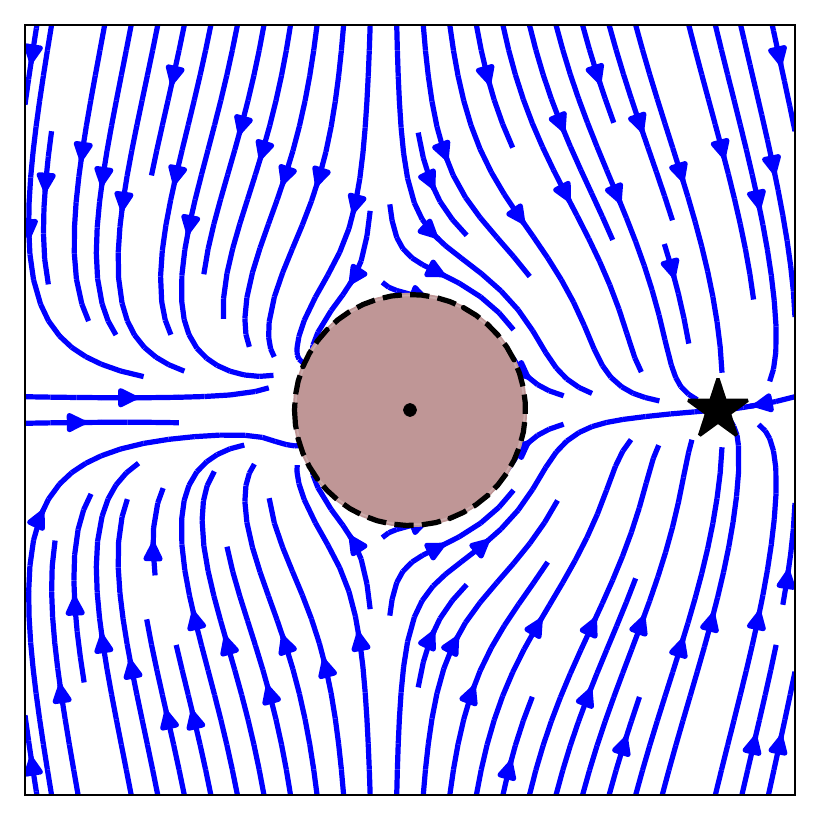}
\caption{Modulated system}
\label{fig:nonlinear_vectorfield_modulated}
\end{subfigure}%
\begin{subfigure}{.33\columnwidth}
\centering
\includegraphics[width=\textwidth]{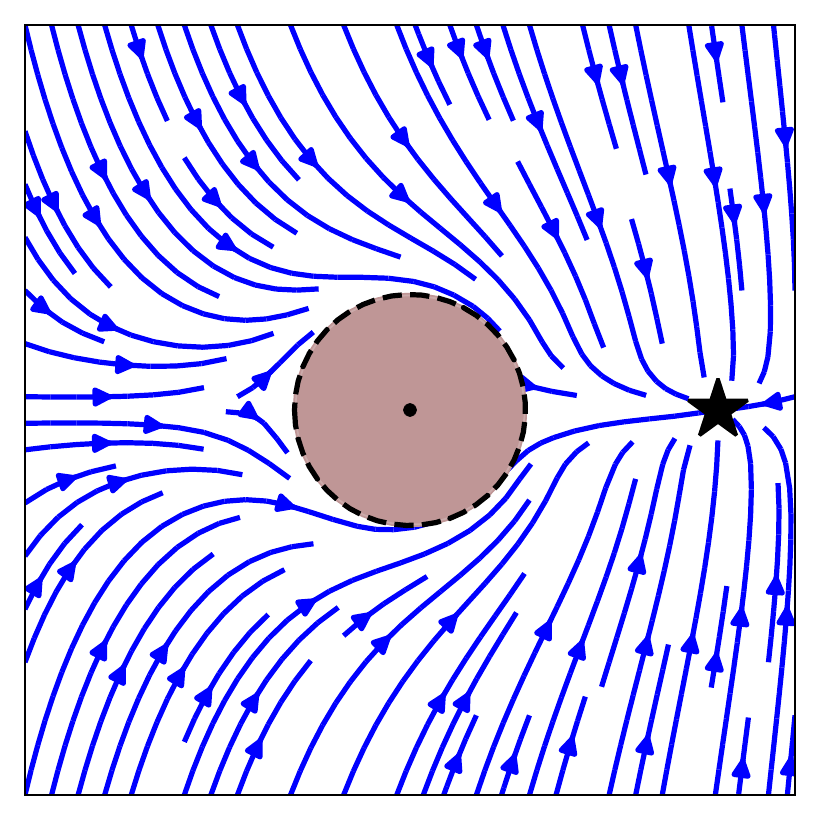}
\caption{Rotated system}
\label{fig:nonlinear_vectorfield_rotated}
\end{subfigure}%
\caption{In surrounding with initially linear dynamics \textbf{(a)}, the basic modulation method can lead to local minima on the surface of the obstacle \textbf{(b)}, while the ROAM ensures only a single saddle point in space \textbf{(c)}.}
\label{fig:comparison_nonlinear_vectorfield}
\end{figure}

The algorithm shows increased convergence for nonlinear dynamical systems (Fig.~\ref{fig:comparison_nonlinear_vectorfield}).

%% file: input/kmeans_obstacle.tex
\section{K-Means Clustering for Avoidance Regions}
Often trajectories can have complicated patterns which are mostly position-dependent. Local motions vary vastly from different regions, and hence finding an Lyapunov function that fits all the space is often not practical.
We introduce a novel method, which divides the space into local clusters, for which we estimate a simple, local Lyapunov function, and then the local (rotation) of the initial velocity is learned.
Finally, the transition between the clusters is ensured by using (nonlinear) inverted obstacle avoidance.

\subsection{Learning the Clustering of the Trajectory}
For the clustering of space, the algorithm uses the K-Means algorithm as implemented in \cite{scikit-learn}.

\subsubsection{Cluster Initialization}
Firstly, the position data is normalized using its variance and mean, i.e. \\
$\hat{\xi}^s = (\xi^s - \xi^{\mathrm{mean}}) / \xi^{\mathrm{var}}$. 

From the velocity, only the direction is important for the creation of the dynamical system. Hence convert the velocities to unit vectors: \\
$\hat{\vect v}^s = \vect v^s / \| \vect v^s \|, \; \forall \, \vect v^s , \; \| \vect v^s \| > 0$

Furthermore, since we use a consecutive trajectory as an input, we keep the sequence-value:
$s = i^s / N^{s}$, with $i^s$ the index of the point in the specific demonstration, and $N^s$ the total number of data points of the specific demonstration.

Finally, the initial clustering is applied to the data, i.e., \\
$\matr X~=~\left[ \hat{\xi}^s \;\; \hat{v}^s \;\; s \right]$.

We assume that for the run-time prediction, i.e., to choose in which cluster a current state is in, we only have access to the current position. (As velocity and sequence value might be wrongly estimated in human-robot interaction.)

An improved clustering distribution can be obtained by using the physical consistent distribution learning introduced by \cite{figueroa2018physically}.



\subsection{Local Dynamics and Cluster Hierarchy}
From Sec.~\ref{sec:rotational_avoidance}, we know that full convergence towards an attractor can be obtained for asymptotically stable, nonlinear dynamical systems with an attractor.


The iterative K-Means clustering and hierarchy evaluation, ensures that the magnitude of the deviation between the reference direction $\vect v^c$ and the rotations stay small, i.e., smaller than $\pi / 2$.

\begin{figure}[t]\centering
\begin{subfigure}{.49\columnwidth}
\centering
\includegraphics[width=\textwidth]{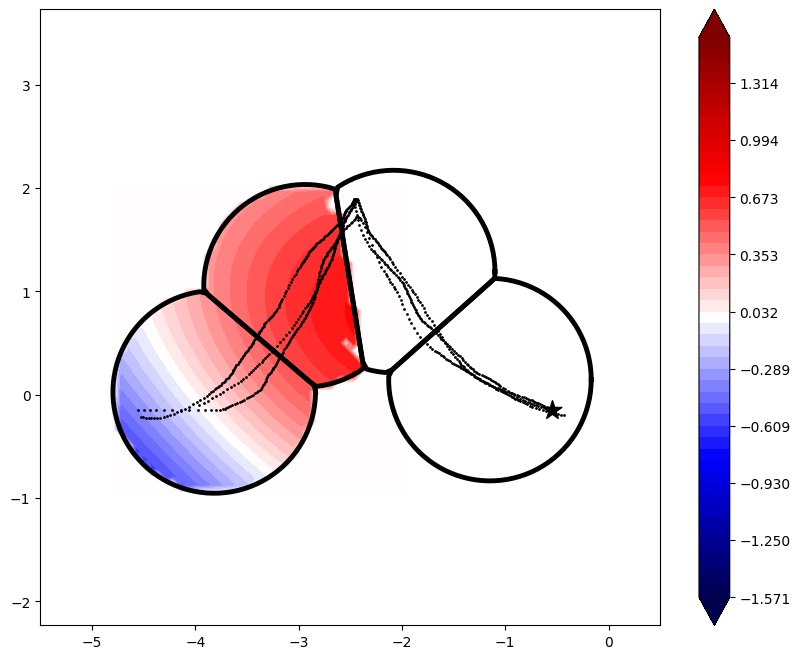}
\caption{Dynamics 1}
\label{fig:dynamics1}
\end{subfigure}%
\begin{subfigure}{.49\columnwidth}
\centering
\includegraphics[width=\textwidth]{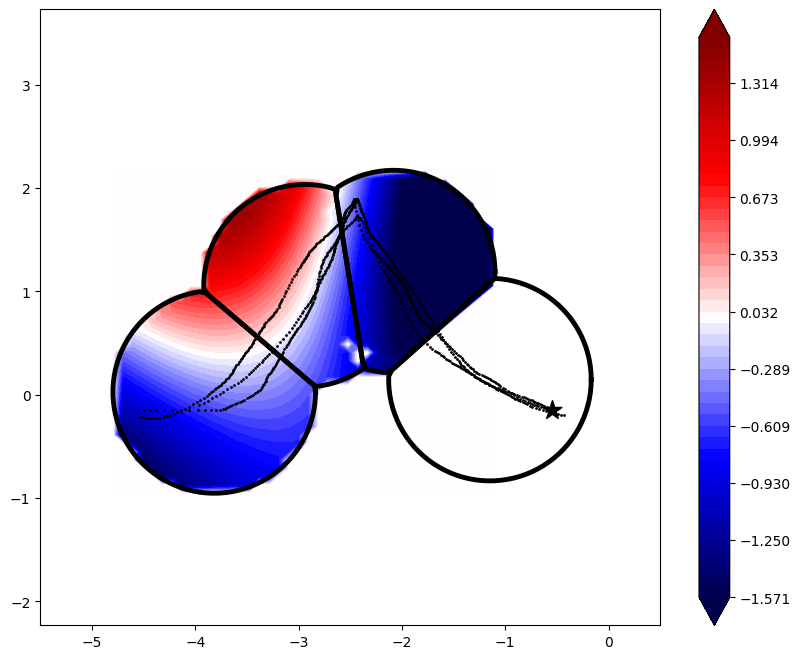}
\caption{Dynamics 2}
\label{fig:dynamics2}
\end{subfigure}
\begin{subfigure}{.49\columnwidth}
\centering
\includegraphics[width=\textwidth]{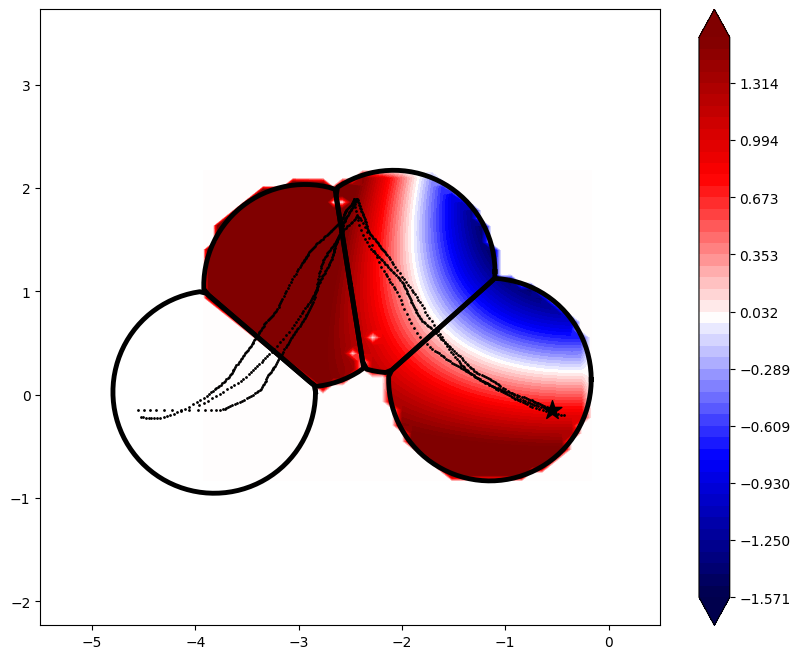}
\caption{Dynamics 3}
\label{fig:dynamics234}
\end{subfigure}%
\begin{subfigure}{.49\columnwidth}
\centering
\includegraphics[width=\textwidth]{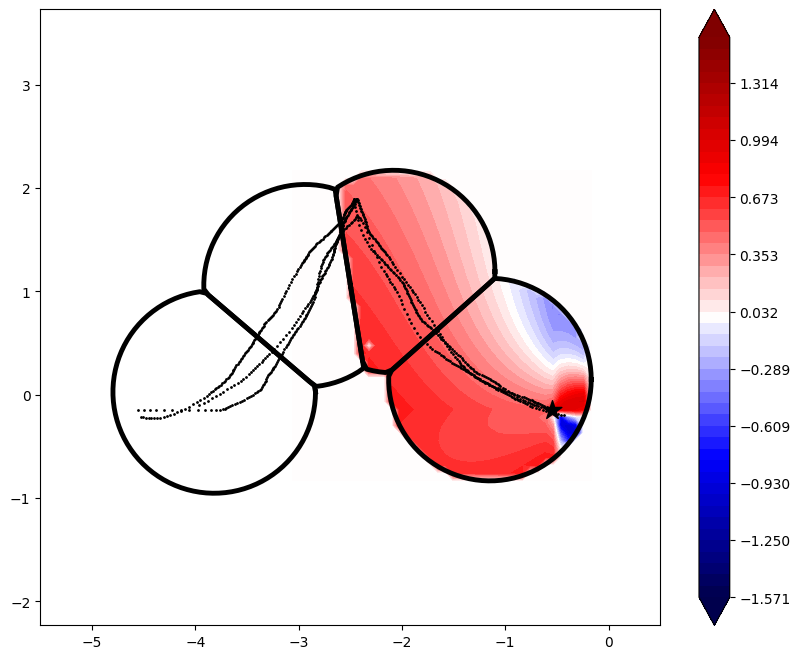}
\caption{Dynamics 4}
\label{fig:dynamics4}
\end{subfigure}%
\caption{The local deviation with respect to the initial dynamics is based on a different reference region. Here, we use here SVR, see \cite{scikit-learn}.}
\label{fig:local_deviation}
\end{figure}

Any common regression technique can be used (see Fig.~\ref{fig:local_deviation})
A regression to obtain the desired deviation can hence be obtained, note that the predicted output is the deviation of the original dynamics $\vect{y}^{\mathrm{dev}}$
\begin{equation}
  y^{\mathrm{dev}} = f^{\mathrm{regr}}(X)
\end{equation}
Note that the regression output $y^{\mathrm{dev}}$ is clipped at a maximal deviation  $\phi^{\mathrm{max}} < \pi / 2$, we choose $\phi^{\mathrm{max}} = 0.4 \pi$.

\begin{figure}[h]
    \centering 
    \includegraphics[width=0.8\columnwidth]{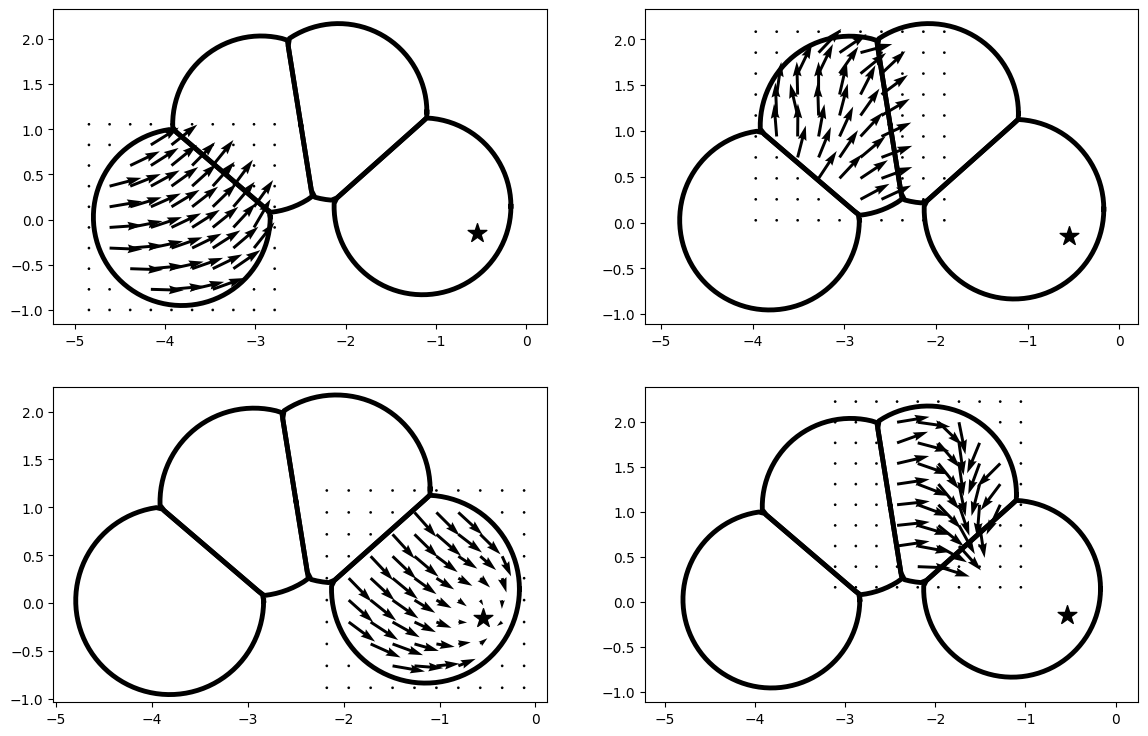}
    \caption{Local deviation with obstacle (hull) avoidance. The transition between the obstacles is ensured by applying the non-linear obstacle avoidance for hulls to the learned dynamics as described in Sec~\ref{sec:rotational_avoidance}). }
    \label{fig:local_deviation_and_avoidance}
\end{figure}




\subsection{K-Means Boundary Obstacle}
From \cite{huber2019avoidance, huber2022avoiding}, we know that for successful obstacle and boundary avoidance, only the reference direction $\vect r(\xi)$, normal $\vect n(\xi)$, and distance value $\Gamma(\xi)$ are needed.

\subsubsection{Distance Value}
We interpret each K-Means cluster as a boundary obstacle. For this we need to evaluate the $\Gamma$ value and normal direction $\vect n(\xi)$ at each position.
First, we define the projected position outside the obstacle, 
\begin{equation}
  \hat{\xi} = 
  \begin{cases}
    \xi & \text{if} \;  \| \tilde \xi \| \geq \| \tilde \xi^b \| \\
    \xi^c + \tilde \xi ( \| \tilde \xi^b \| / \| \tilde \xi \|)^2  & \text{otherwise}
  \end{cases}
\end{equation}
with $\tilde \xi = \xi - \xi^c$ and $\tilde \xi^b = \xi^b - \xi^c$ being the relative positions. The position $\hat \xi$ will be used for further calculation.

The boundary point $\xi^b$ defines the separation between the obstacles and are defined as $\xi^b = \xi + b \vect r(\xi)$, and $\Gamma(\xi^b) = 1$.
The distance to the boundary is the shortest with respect to all clusters, i.e.,
\begin{equation}
  b = \min_{i \neq o} \{ b_i(\xi) : b_i(\xi) > 0 \} \label{eq:boundary_factor}
\end{equation}

The intersection distance between to the cluster $i$
\begin{equation}
b_{i}(\xi) =
\begin{bmatrix}
  -\vect{r}(\xi) & \matr B^{\vect n}_{[1 .. D ]}
\end{bmatrix} ^{-1} 
\frac{1}{2} \left(  \xi^c_i - \xi^c  \right)
\end{equation}
with $D$ the dimension and $\matr B$ the orthogonal basis to the normal vector of the separating plane $(\xi^c_i - \xi) / \| \xi^c_i - \xi \|$.

The distance value $\Gamma(\xi)$ is defined as
\begin{equation}
  \Gamma(\xi) =  \sum_i w^\Gamma_i(\xi) r^l_i(\xi) / \| \xi - \xi^c \|
  \; ,  \;\;
  r^l_i(\xi) = \| \hat{\xi} \|  - d_i^n(\xi)
  \label{eq:gamm_kmeans}
\end{equation}
The normal distances are:
\begin{equation}
  d_i^n(\xi) = \dotprod{\frac{\xi^c_i - \xi^c}{ \| \xi^c_i - \xi^c \| }}{\xi - \frac{\xi^c_i + \xi^c}{2}}
\end{equation}

and the weights
\begin{equation*}
  w^\Gamma_i(\xi) = w^0_i(\xi) / {\sum_i w^0_i(\xi)}
  \;\; \text{with} \quad
  w^0_i(\xi) = \max \left( d_i^n(\xi), \; 0 \right)
\end{equation*}

Since the dynamical system needs to be able to traverse the boundary, the distance function has to be large at the \textit{transparent} boundary and not deflect the motion, i.e., $\Gamma \gg 1$,
let us assume that a wall $t$ is transparent, in this case following is adapted:
\begin{gather*}
  \hat{w}^\Gamma_i(\xi) = \max( w_i(\xi) - w_t(\xi), \, 0) \; \forall \, i \notin \{t, o\} 
  \\
  \hat{w}^\Gamma_t(\xi) = 1 - \sum_{i \neq t} \hat{w}^\Gamma_i(\xi)
\end{gather*}

and additionally, the radius of the local radius of the transparent wall $t$ as given in \eqref{eq:gamm_kmeans} is set as:
\begin{equation}
  r^l_t(\xi) = \left( \| \hat{\xi} \|  - d_i^n(\xi) \right) / \left( 1 - w^\Gamma_t(\xi) \right)
\end{equation}

\subsubsection{Normal Direction}
The weights of the normal directions are similar, and are given by:
\begin{equation}
  w^n_i = 
  \begin{cases}
    0 & \text{if} \;\; i = o \\
    w^\Gamma_i & \text{otherwise}  
  \end{cases}
\end{equation}

  
To obtain the final normal direction $\vect n(\xi)$, the directional weighted sum is taken as described in \cite{huber2022avoiding}, with the above described weights $w^n_i$ and null direction $-\vect r(\xi)$.

Note, that the normal direction does not take into consideration the transition-boundaries. Since this is already taken into account by the $\Gamma$-value. The local avoidance can be seen in Fig.~\ref{fig:local_deviation_and_avoidance}.

\begin{figure}[t]\centering
\begin{subfigure}{.49\columnwidth}
\centering
\includegraphics[width=\textwidth]{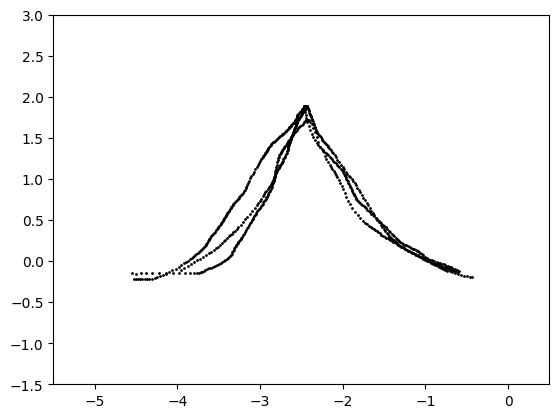}
\caption{Initial Data}
\label{fig:a_shape_data}
\end{subfigure}%
\begin{subfigure}{.49\columnwidth}
\centering
\includegraphics[width=\textwidth]{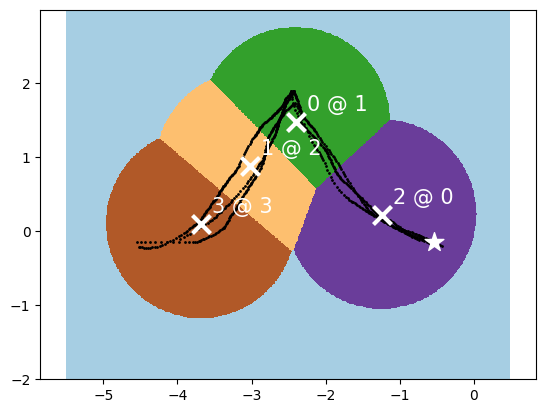}
\caption{Cluster in position, direction, and sequence.}
\label{fig:a_shape_kmeans}
\end{subfigure}
\begin{subfigure}{.49\columnwidth}
\centering
\includegraphics[width=\textwidth]{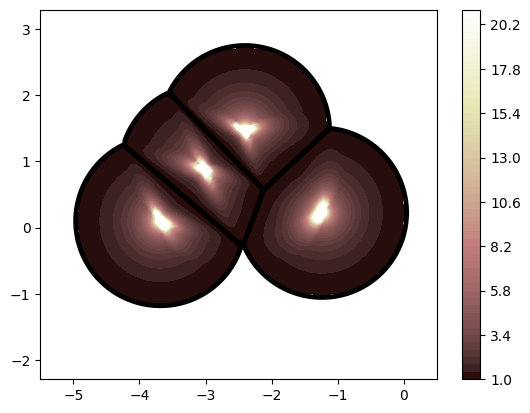}
\caption{Gamma-values of the sub-clusters}
\label{fig:a_shape_gamma}
\end{subfigure}%
\begin{subfigure}{.49\columnwidth}
\centering
\includegraphics[width=\textwidth]{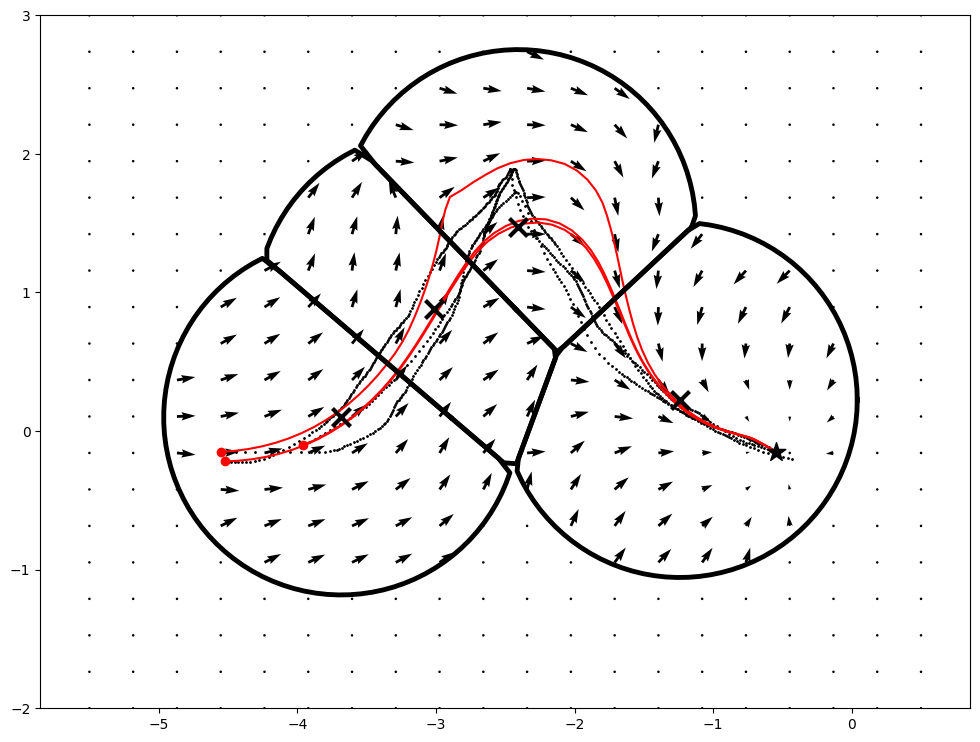}
\caption{Final trajectory prediction.}
\label{fig:a_shape_learned}
\end{subfigure}%
\caption{The prediction uses the proposed cluster-regression algorithm for stable motion learning applied to the A-shape in the LASA-dataset \cite{khansari2014learning}.}
\label{fig:a_shape_learning}
\end{figure}




%% file: input/conclusion.tex
\section{Conclusion}
Representing the dynamics as a local rotation has shown good results. Both for obstacle avoidance, but also motion learning (see Fig.~\ref{fig:a_shape_learning}).
Current algorithms are in their infant state but show promising results. Future work will include convergence proof for all algorithms as well as combining motion learning and obstacle avoidance which can ensure stability guarantees from the previous.